\documentstyle[aclap]{article}
\title{\vspace{-0.5in}An Empirical Approach to Temporal Reference Resolution}
\author{Janyce Wiebe, Tom O'Hara,  Kenneth McKeever, and 
Thorsten~\"{O}hrstr\"{o}m-Sandgren \\
Dept. of Computer Science and the Computing Research Laboratory \\
New Mexico State University  \\
Las Cruces, NM 88003    \\
{\tt wiebe,tomohara,kmckeeve,tsandgre@cs.nmsu.edu}} 

\begin{document}
\maketitle
\vspace{-0.5in}
\begin{abstract}
This paper presents the results of an empirical investigation of temporal
reference resolution in scheduling dialogs.  
The algorithm adopted is
primarily a linear-recency based approach that does not include a
model of global focus.  
A fully automatic system has been developed
and evaluated on unseen test data with good results.  This paper 
presents the results of an intercoder reliability study, a
model of temporal reference resolution that supports linear recency
and has very good coverage, the results of the system evaluated
on unseen test data,
and a detailed analysis of the dialogs assessing the viability
of the approach. 
\end{abstract}
\section{Introduction}
Temporal information is often a significant part of the
meaning communicated in dialogs and texts, but is often 
left implicit, to be recovered by the listener or
reader from the surrounding context.  Determining all of
the temporal information that is being conveyed can be
important for many interpretation tasks.
For instance, in machine translation, knowing the temporal
context is important in translating sentences with missing 
information. This
is particularly useful when dealing with noisy data, as with spoken 
input (Levin et al. 1995). In
the following example, the third utterance could be interpreted in three
different ways.
\begin{quote}
\begin{tabular}{ll} 
s1: & (Ahora son las once y diez) \\ 
    & Now it is eleven ten\\ 
s1: & (Qu\'{e} tal a las doce) \\ 
    & How about twelve \\
s1: & (Doce a dos) \\
    & Twelve to two \\
{\em or} & The twelfth to the second \\
{\em or} & The twelfth at two \\
\end{tabular}
\end{quote}
By maintaining the temporal context (i.e., the 5th of March 1993 at
12:00),
the system will know that ``12:00 to 2:00'' is a more probable 
interpretation than ``the 12th at 2:00''.

In addition, maintaining the temporal context would be useful for
information extraction tasks dealing with natural language texts such
as memos or meeting notes.  For instance, it can be used to resolve
relative time expressions, so that absolute dates can be entered in 
a database with a uniform representation.

This paper presents the results of an empirical investigation of
temporal reference resolution in scheduling dialogs (i.e., dialogs in
which participants schedule a meeting with one another).  This work thus
describes how to identify temporal information that is missing due to
ellipsis or anaphora, and it shows how to determine the times evoked
by deictic expressions.
In developing the algorithm, our approach was to start with
a straightforward, recency-based approach and add complexity as needed
to address problems encountered in the data.  The algorithm does not
include a mechanism for handling global focus (Grosz \& Sidner 1986),
for centering within a discourse segment (Sidner 1979; Grosz et
al. 1995), or for performing tense and aspect interpretation.
Instead, the algorithm processes anaphoric references with respect to
an Attentional State (Grosz \& Sidner 1986) structured as a linear
list of all times mentioned so far in the current dialog.  The list is
ordered by recency, no entries are ever deleted from the list, and
there is no restriction on access.  The algorithm decides among
candidate antecedents based on a combined score reflecting recency, a
priori preferences for the type of anaphoric relation(s) established,
and plausibility of the resulting temporal reference.  In determining
the candidates from which to choose the antecedent, for each type of
anaphoric relation, the algorithm considers only the most recent
antecedent for which that relationship can be established.

The algorithm was primarily developed on a corpus of Spanish dialogs
collected under the JANUS project (Shum et al. 1994) (referred to
hereafter as the ``CMU dialogs'') and has also been applied to a
corpus of Spanish dialogs collected under the Artwork project (Wiebe
et al. 1996) (hereafter referred to as the ``NMSU dialogs'').
In both cases, subjects were told that they were to set up a meeting based on
schedules given to them detailing their commitments.  The CMU protocol
is akin to a phone conversation between people who do not know each
other.  Such strongly task-oriented dialogs would arise in many useful
applications, such as automated information providers and automated
phone operators.  The NMSU data are face-to-face dialogs between
people who know each other well.  These dialogs are also strongly
task-oriented, but only in these, not in the CMU dialogs, do the
participants stray significantly from the scheduling task.  In
addition, the data sets are challenging in that they both include
negotiation, both  contain many disfluencies, and both show
a great deal of variation in how dates and times are discussed.

To support the computational work,  
the temporal references in the corpus were manually
annotated according to explicit coding instructions.  
In addition, we annotated the seen training dialogs for 
anaphoric chains, to support analysis of the data.

A fully automatic system has been developed that takes as input the
ambiguous output of a semantic parser (Lavie \& Tomita 1993, Levin et
al. 1995).  The system performance on unseen, held-out test data is
good, especially on the CMU data, showing the usefulness of our
straightforward approach.  The performance on the NMSU data is worse
but surprisingly comparable, given the greater complexity of the data
and the fact that the system was primarily developed on the simpler
data.

Ros\'{e} et al. (1995), Alexandersson et al. (1997), and Busemann et
al. (1997) describe other recent NLP systems that resolve temporal
expressions in scheduling dialogs as part of their overall
processing, but they do not give results of system performance
on any temporal interpretation tasks.
Kamp \& Reyle (1993) address many
representational and processing issues in the interpretation of
temporal expressions, but they do not attempt coverage of a data set
or present results of a working system.  To our knowledge, there are
no other published results on unseen test data of systems performing
the same temporal resolution tasks.

The specific contributions of this paper are the following.
The results of an intercoder reliability study involving 
naive subjects are presented (in section
\ref{intercoder}) as well as
an abstract presentation of a
model of
temporal reference resolution (in section
\ref{model}).   In addition, the
high-level algorithm is
given (in section \ref{algorithm}); the fully refined algorithm,
which distinguishes many more subcases than can be presented here,
is available online at 
http://crl.nmsu.edu/Research/Projects/artwork.
Detailed results of an implemented system are also presented
(in section \ref{results}),
showing the success of the algorithm.
In the final part of the paper, we
abstract away from matters of implementation and analyze the
challenges presented by the dialogs to an algorithm
that does not include a model of global focus (in section
\ref{global}).  We found surprisingly few such challenges.

\section{The Corpus and Intercoder Reliability Study}
\label{intercoder}
Consider this passage from the corpus (translated into English):

\noindent
\vspace*{2mm}
\begin{tabular}{lll}
\multicolumn{3}{l}{{\bf Preceding time:}
Thursday 19 August} \\
s1& 1&   On Thursday I can only meet after two pm \\
   &2&  From two to four \\
   &3&     Or two thirty to four thirty \\
   &4&    Or three to five \\
s2& 5&   Then how does from two thirty to  \\
  &  & four thirty seem to you \\
   &6&   On Thursday \\
s1& 7&  Thursday the thirtieth of September \\
\end{tabular}

\vspace*{4mm}
An example of temporal reference resolution is that
(2) refers to 2-4pm Thursday 19 August.
Although related, this problem is distinct from
tense and aspect interpretation in discourse (as addressed in, e.g., 
Webber 1988,
Song \& Cohen 1991, Hwang \& Schubert 1992, Lascarides et al. 1992,
and Kameyama et al. 1993).

Because the dialogs are centrally concerned with negotiating
an interval of time in which to hold a meeting, our
representations are geared toward such intervals.
Our basic representational unit is given in figure \ref{tempunit}. 
To avoid confusion, we refer to this basic unit throughout 
as a {\it Temporal Unit} ({\it TU}).  
\begin{figure*}
\begin{center}
\vspace*{2mm}
\begin{tabular}{|lllll|}
\hline
((start-month, & start-date, & start-day-of-week, & start-hour\&minute,
& start-time-of-day) \\
\hspace*{1mm}(end-month, & end-date, & end-day-of-week, & end-hour\&minute, &
end-time-of-day))  \\
\hline
\end{tabular}
\end{center}
\caption{Temporal Units}
\label{tempunit}
\end{figure*}
\noindent

The time referred to in, for example, 
``From 2 to 4, on Wednesday the 19th of August'' is represented as: 
\begin{quotation}

\noindent
((August, 19th, Wednesday, 2, pm) \\
\hspace*{1mm}(August, 19th, Wednesday, 4, pm))
\end{quotation}
Thus, the information from multiple noun phrases is often
merged into a single representation of the underlying interval
evoked by the utterance.  

An utterance such as ``The meeting starts at 2'' is represented as an
interval rather than as a point in time, reflecting the orientation of
the coding scheme toward intervals.  Another issue this kind of 
utterance raises is whether or not a speculated ending time
of the interval should be filled in, using knowledge of how
long meetings usually last.  In the CMU data, the meetings
all last two hours.  However,
so that the instructions will
be applicable to a wider class of dialogs, we decided to
be conservative with respect to filling in an ending time,
given the starting time (or vice versa), leaving it open unless
something in the dialog explicitly suggests otherwise. 

There are cases in which 
times are considered as points
(e.g., ``It is now 3pm'').
These are 
represented as Temporal Units with the same starting and ending times (as in
Allen (1984)).  
If just one ending point is represented, all the fields of the
other are null.  And, of course, all fields are null for
utterances that do not contain temporal information.
In the case of an utterance that refers to multiple, distinct
intervals, the representation is a list of Temporal Units.

A Temporal Unit is also the representation used in the evaluation 
of the system.
That is, the  system's answers
are mapped from its more complex internal representation (an
{\it ILT}, see section \ref{architecture}) into this
simpler vector representation before evaluation is performed.

As in much recent empirical work in discourse processing (e.g.,
Arhenberg et al. 1995; Isard \& Carletta 1995; Litman \& Passonneau
1995; Moser \& Moore 1995; Hirschberg \& Nakatani 1996), we performed
an intercoder reliability study investigating agreement in annotating
the times.  The goal in developing the annotation
instructions is that they can be used reliably by non-experts after a
reasonable amount of training (cf. Passonneau \& Litman 1993, Condon
\& Cech 1995, and Hirschberg \& Nakatani 1996), where reliability is
measured in terms of the amount of agreement among annotators.
High reliability indicates that the encoding scheme is reproducible
given multiple labelers.
In addition, the instructions serve to document the
annotations. 

The subjects were three people with no previous 
involvement in the project. They were given the original
Spanish and the English translations. However, as they have limited knowledge
of Spanish, in essence they annotated the English translations.

The subjects annotated two training dialogs according to the instructions.
After receiving feedback, they annotated four unseen test dialogs.  
Intercoder reliability was assessed using Cohen's Kappa
statistic ($\kappa$) 
(Siegel \& Castellan 1988, Carletta 1996).

$\kappa$ is calculated as follows,
where the numerator is the average
percentage agreement among the annotators (Pa) 
less a term for chance agreement (Pe), and the 
denominator is 
100\% agreement less the same term for chance agreement (Pe):

\[ \kappa = \frac{Pa - Pe}{1 - Pe} \]

\noindent
(For details on calculating Pa and Pe see Siegel \& Castellan
1988).  
As discussed in (Hays 1988), $\kappa$ will be 0.0 when the agreement
is what one would expect under independence, 
and it will be 1.0 when the agreement
is exact.
A $\kappa$ value of 0.8 or greater indicates a high level of reliability
among raters, with values between 0.67 and 0.8 indicating only moderate
agreement (Hirschberg \& Nakatani 1996; Carletta 1996).

In addition to measuring intercoder reliability, we compared each
coder's annotations to the evaluation Temporal Units 
used to assess the system's
performance.  
These evaluation Temporal Units were assigned by an expert working
on the project.

The agreement among coders ($\kappa$) is shown in table
\ref{kappa-stats}. In addition, this table shows the average pairwise
agreement of the coders and the expert ($\kappa_{avg}$), which was
assessed by averaging the individual $\kappa$ scores (not shown).
\begin{table*}
\begin{center}
\begin{tabular}{|l|l|l|c||c|}
\hline
Field        &      Pa &   Pe & $\kappa$ & $\kappa_{avg}$ \\
\hline					   
start        &         &      &          &     \\
Month        &     .96 &  .51 &  .93     & .94 \\
Date         &     .95 &  .50 &  .91     & .93 \\
WeekDay      &     .96 &  .52 &  .91     & .92 \\
HourMin      &     .98 &  .82 &  .89     & .92 \\
TimeDay      &     .97 &  .74 &  .87     & .74 \\
\hline					   
end          &         &      &          &     \\
Month        &     .97 &  .51 &  .93     & .94 \\
Date         &     .96 &  .50 &  .92     & .94 \\
WeekDay      &     .96 &  .52 &  .92     & .92 \\
HourMin      &     .99 &  .89 &  .90     & .88 \\
TimeDay      &     .95 &  .85 &  .65     & .52 \\
\hline
\end{tabular}
\caption{Agreement among Coders (kappa coefficients by field)}
\label{kappa-stats}
\end{center}
\end{table*}
There is a moderate or high level of agreement among
annotators in all cases except the ending time of day, a weakness we are
investigating.  Similarly, there are reasonable levels of agreement
between our evaluation Temporal Units and the answers the naive coders
provided.

Busemann et al. (1997) also annotate temporal information in a corpus
of scheduling dialogs.  However, their annotations are at the level of
individual expressions rather than at the level of Temporal Units, and
they do not present the results of an intercoder reliability study.

\section{Model}
\label{model}
This section presents our model of temporal reference in scheduling
dialogs.  The treatment of anaphora in this paper is as a relationship
between a Temporal Unit representing a time evoked in the current
utterance, and one representing a time evoked in a previous utterance.
The resolution of the anaphor is a new Temporal Unit that represents
the interpretation of the contributing words of the current utterance.

\begin{figure*}
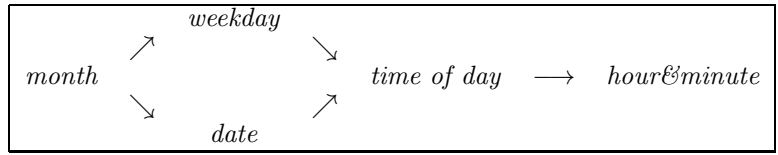

\begin{center}
\def\ne{$\nearrow$}
\def\se{$\searrow$}
\def\e{$\longrightarrow$}
\noindent
{\em
\begin{tabular}{|lrclccc|}
\hline
      &     & weekday &      &             &    &               \\
      & \ne &         & \se  &             &    &               \\
month &     &         &      & time of day & \e & hour\&minute  \\
      & \se &         & \ne  &             &    &               \\
      &     & date    &      &             &    &               \\
\hline
\end{tabular}
}
\end{center}
\caption{Specificity Ordering}
\label{specificity}
\end{figure*}

Fields of Temporal Units are partially ordered as in figure \ref{specificity},
from least to most specific.

In all cases below, after the resolvent has been formed, it
is subjected to highly accurate, trivial inference to
produce the final interpretation
(e.g., filling in the day of the week given the month and the date). \\

\noindent
\underline{The cases of non-anaphoric reference:}
\begin{enumerate}
\item 
A deictic expression is resolved into a time interpreted with 
respect to the dialog date (e.g., ``Tomorrow'', ``last week'').
{
(See rule NA1 in section \ref{rules}.)}
\item
\label{fordeictic}

A forward time is calculated by using the dialog date as a frame of reference. \\
Let $F$ be the most specific field in $TU_{current}$
above the level of {\it time-of-day}.  \\
Resolvent: The next $F$ after the dialog date, augmented with
the fillers of the fields in $TU_{current}$ at or below the level of 
{\it time-of-day}. 
{ 
(See rule NA2.)}  \\

For both this and anaphoric
relation (\ref{forana}), 
there are subcases for whether the starting and/or ending
times are involved.  
Note that tense can influence the choice of whether to calculate a forward
or a backward time from a frame of reference
(Kamp \& Reyle 1993), but we do not account for this in our
model due to the lack of tense variation in the corpora.

\vspace*{2mm}
\begin{tabular}{ll}
Ex: & Dialog date is Mon, 19th, Aug  \\
    & ``How about Wednesday at 2?'' \\
    &   \hspace*{3mm} interpreted as 2 pm, Wed 21 Aug\\
\end{tabular}

\vspace*{2mm}
\end{enumerate}

\noindent
\underline{The cases of anaphora considered:}
\begin{enumerate}
\item
The utterances evoke the same time, or the second is more specific
than the first. \\
Resolvent: the union of the information in the two Temporal Units.
{ 
(See rule A1.)}

\noindent

\vspace*{2mm}
\begin{tabular}{ll}
Ex: & ``How is Tuesday, January 30th?'' \\
    & ``How about 2?''\\
\end{tabular}

(See also (1)-(2) of the corpus example.) 

\item
The current utterance evokes a time that 
includes the time evoked by a previous time,
and the current time is less specific.
{
(See rule A2.)}

Let $F$ be the most specific field in $TU_{current}$. \\
Resolvent:  All of the information in $TU_{previous}$ from $F$ on up. 

\begin{tabular}{ll}
Ex: & ``How about Monday at 2?'' \\
    &   \hspace*{3mm} resolved to 2pm, Mon 19 Aug  \\
    & ``Ok, well, Monday sounds good.'' \\
\end{tabular}

(See also (5)-(6) in the corpus example.)
\item
\label{forana}
This is the same as non-anaphoric case (\ref{fordeictic}) above,
but the new time is calculated with respect to $TU_{previous}$
instead of the dialog date. 
{
(See rule A3.)}

\begin{tabular}{ll}
Ex: & ``How about the 3rd week in August?'' \\
    & ``Let's see, Monday sounds good.''  \\
    &   \hspace*{3mm} interpreted as Mon, 3rd week in Aug \\
\end{tabular}

\vspace*{2mm}
\begin{tabular}{ll}
Ex: & ``Would you like to meet Wed, Aug 2nd?'' \\
    & ``No, how about Friday at 2.''  \\
    &   \hspace*{3mm} interpreted as Fri, Aug 4 at 2pm\\
\end{tabular}

\item
\label{revise}
The current time is a modification of the previous time; the times
are consistent down to some level of specificity $X$ and differ
in the filler of $X$. \\
Resolvent:  The information in $TU_{previous}$ above level $X$ 
together with the information in $TU_{current}$ at and below level $X$. 
{
(See rule A4.)}

\begin{tabular}{ll}
Ex: & ``Monday looks good.'' \\
    &  \hspace*{3mm} resolved to Mon 19 Aug \\
    &``How about 2?'' \\
    &   \hspace*{3mm} resolved to 2pm Mon 19 Aug \\
    & ``Hmm, how about 4?''  \\
    &   \hspace*{3mm} resolved to 4pm Mon 19 Aug \\
\end{tabular}

\noindent
(See also (3)-(5) in the example from the corpus.)
\end{enumerate}

Although we found domain knowledge and task-specific
linguistic conventions most useful, 
we observed in the NMSU data some instances of
potentially exploitable syntactic information to pursue
in future work (Grosz et al. 1995, Sidner 1979).  
For example, ``until'' in the following 
suggests that the first utterance specifies an ending time.
\begin{quotation}
\noindent
``... could it be until around twelve?'' \\
``12:30 there''
\end{quotation}
A preference for parallel syntactic roles
might be used to recognize that the second utterance specifies an
ending time too.
\section{The Algorithm}
\label{algorithm}
This section presents our algorithm for temporal reference
resolution.  After a brief overview, the rule-application
architecture is described and then the rules composing
the algorithm are given. 
As mentioned earlier, this is a high-level algorithm.
Description of 
the complete algorithm, including a specification of the
normalized input representation (see section \ref{architecture}), can be 
obtained from a report available at the project web page
(http://crl.nmsu.edu/Research/Projects/artwork).

There is a rule for each of the relations presented
in section \ref{model}.
Those for the anaphoric relations involve various applicability
conditions on the current utterance and a potential antecedent.
For the
current not-yet-resolved Temporal Unit, each rule is applied.  For
the anaphoric rules, the antecedent
considered is the most recent one meeting the conditions.
All consistent
maximal mergings of the results are formed, and the one with the
highest score is the chosen interpretation.

\subsection{Architecture}
\label{architecture}
Following (Qu et al. 1996) and (Shum et al. 1994), the   
representation of a single utterance is called an {\it ILT} (for
InterLingual Text).  An ILT, once it has been augmented by our system
with temporal (and speech-act) information, is called an 
{\it augmented ILT} (an {\it AILT}).
The input to our system, produced by a semantic parser
(Shum et al. 1994; Lavie \& Tomita 1993), 
consists of multiple alternative ILT
representations of utterances.
To produce one ILT, 
the parser maps the main event and its participants into
one of a small set of case frames (for example, a {\it meet} frame or
an {\it is busy} frame) and produces a surface representation of any
temporal information, which is faithful to the input utterance.
Although the events and states discussed in the NMSU
data are often outside the coverage of this parser, the
temporal information generally is not. Thus, the parser provides us
with a sufficient input representation for our purposes on both sets
of data.  This parser is proprietary, but it would not be difficult to
produce just the portion of the temporal information that our system
requires.

Because the input consists of alternative sequences of ILTs, the
system resolves the ambiguity in batches.  In
particular, for each input sequence of ILTs, it produces a sequence of
AILTs and then chooses the best sequence for the corresponding
utterances.  In this way, the input ambiguity is resolved as a
function of finding the best temporal interpretations of the utterance
sequences in context (as suggested in Qu et al. 1996).

A focus list keeps track of what has been discussed so far in the
dialog.  After a final AILT has been created for the current
utterance, the AILT and the utterance are placed together on the focus
list (where they are now referred to as a {\it discourse entity}, or
{\it DE}).  In the case of utterances that evoke more than one
Temporal Unit, 
a separate entity is added for each to the focus list
in order of mention.  

Otherwise, the system architecture is similar
to a standard production system, with one major exception: rather than
choosing the results of just one of the rules that fires (i.e.,
conflict resolution), multiple results can be merged.  This is a
flexible architecture that accommodates sets of rules targeting
different aspects of interpretation, allowing the system to take
advantage of constraints that exist between them (for example,
temporal and speech act rules). \\

\noindent
{\bf Step 1.}  The input ILT is {\it normalized}.  In the input ILT, different
pieces of information about the same time might be represented
separately in order to capture relationships among clauses.  Our
system needs to know which pieces of information are about the same
time (but does not need to know about the additional relationships).
Thus, we map from the input representation into a normalized
form that shields the reasoning component from the idiosyncracies of
the input representation.  
After the normalization process, highly accurate, obvious inferences are
made and added to the representation.  \\

\noindent
{\bf Step 2.}
All rules are applied to the normalized input.  
The result of a rule
application is a {\it partial AILT} ({\it PAILT})---information this
rule would contribute to the interpretation of the utterance.  This
information includes a certainty factor representing an a priori
preference for the type of anaphoric or non-anaphoric relation being
established. In the case of anaphoric relations, this factor gets
adjusted by a term representing how far back on the focus list the
antecedent is (in rules A1-A4 in section \ref{rules}, the adjustment
is represented by {\it distance factor} in the calculation of the
certainty factor {\it CF}).
The result of this step is the set of PAILTs produced
by the rules that fired (i.e., those that succeeded). \\

\noindent
{\bf Step 3.}
All maximal mergings of the PAILTs 
are created.
Consider a graph in which the PAILTs are the vertices, and there is
an edge between two PAILTs iff the two PAILTs are compatible.  Then,
the maximal cliques of the graph (i.e., the maximal complete subgraphs)
correspond to the maximal mergings.
Each maximal merging is then merged with the normalized
input ILT,
resulting in a set of AILTs.  \\

\noindent
{\bf Step 4.}
The AILT chosen is the one with the highest certainty factor.
The certainty factor of an AILT is calculated as follows.
First, the certainty factors of the constituent PAILTs
are summed. Then,
critics are applied
to the resulting AILT, 
lowering the certainty factor if the information is judged
to be incompatible with the dialog state.  

The merging process might have yielded additional
opportunity for making obvious inferences, so that process
is performed again, to produce the final AILT.

\subsection{Temporal Resolution Rules}
\label{rules}
The rules described in this section (see figure \ref{temporal-rules}) 
apply to individual temporal units
and return either a more-fully specified TU or an empty structure to
indicate failure.  

Many of the rules calculate temporal information with respect to a
frame of reference, using a separate calendar utility.
The following describe these and other functions assumed by the rules
below, as well as some conventions used.

\begin{description}
\item[next($TimeValue$, $RF$):] returns the 
next $timeValue$ that follows reference frame $RF$. \\
next(Monday, [$\ldots$Friday, 19th,$\ldots$]) = Monday, 22nd.

\item[resolve\_deictic($DT$, $RF$):] resolves the 
deictic term $DT$ with respect to the reference frame $RF$.

\item[merge($TU1$, $TU2$):]
if temporal units $TU1$ and $TU2$ contain no conflicting field fillers,
returns a temporal unit containing all of the information
in the two; otherwise returns $\{ \}$.

\item[merge\_upper($TU1$, $TU2$):] 
like the previous function, except includes only those field fillers from $TU1$
that are of the same or less specificity 
as the most specific field filler in $TU2$.

\item[specificity($TU$):]
returns the specificity of the most specific field in $TU$.

\item[starting\_fields($TU$):]
returns a list of starting field names for those in $TU$ having 
non-null values.

\item[structure$\rightarrow$component:]
returns the named component of the structure.

\item[conventions:]
Values are in {\bf bold face} and variables are in $italics$.
$TU$ is the current temporal unit being resolved.
$TodaysDate$ is a representation of the dialog date. $FocusList$ is
the list of discourse entities from all previous utterances.
\end{description}

{
\begin{figure*}

\begin{center}
{\large \em Rules for non-anaphoric relations}
\end{center}

\begin{tabbing}
mm\=mm\=mm\=mm\=mm\=mm\=mm\=mm\= \kill
{\em \bf Rule NA1:} All cases of non-anaphoric relation 1. \\
if there is a deictic term, $DT$, in $TU$ then \\
\>    return \{[when, resolve\_deictic($DT$, {\bf $TodaysDate$})], [certainty, 0.9]\} \\
 \\
{\em \bf Rule NA2:} The starting-time cases of non-anaphoric relation 2. \\
if (most\_specific(starting\_fields($TU$)) $<$ {\bf time\_of\_day}) then \\
\> Let $f$ be the most specific field in starting\_fields($TU$) \\
\> return \{[when, next($TU$$\rightarrow$$f$, $TodaysDate$)], [certainty, 0.4]\}
\end{tabbing}

\vspace*{5mm}
\begin{center}
{\large \em Rules for anaphoric relations}
\end{center}

\begin{tabbing}
mm\=mm\=mm\=mm\=mm\=mm\=mm\=mm\= \kill
{\em \bf Rule A1:} All cases of anaphoric relation 1. \\
for each non-empty temporal unit $TU_{fl}$ from $FocusList$ (starting with most recent) \\
\> if specificity($TU_{fl}$) $\leq$ specificity($TU$) 
         and not empty merge($TU_{fl}$, $TU$) then \\
\>  \> $CF$ = 0.8 $-$ distance\_factor($TU_{fl}$, {\bf $FocusList$}) \\
\>  \> return \{[when, merge($TU_{fl}$, $TU$)], [certainty, $CF$]\} \\
 \\

{\em \bf Rule A2:} All cases of anaphoric relation 2. \\
for each non-empty temporal unit $TU_{fl}$ from $FocusList$ (starting with most recent) \\
\> if specificity($TU_{fl}$) $>$ specificity($TU$) 
      and not empty merge\_upper($TU_{fl}$, $TU$) then \\
\> \> $CF$ = 0.5 $-$ distance\_factor($TU_{fl}$, {\bf $FocusList$}) \\
\> \> return \{[when, merge\_upper($TU_{fl}$, $TU$)], [certainty, $CF$]\} \\
 \\

{\em \bf Rule A3:} Starting-time case of anaphoric relation 3. \\
if (most\_specific(starting\_fields($TU$)) $<$ {\bf time\_of\_day}) then \\
\> for each non-empty temporal unit $TU_{fl}$ from $FocusList$ (starting with most recent) \\
\> \> if specificity($TU$) $\geq$ specificity($TU_{fl}$) then \\
\> \> \> Let $f$ be the most specific field in starting\_fields($TU$) \\
\> \> \> $CF$ = 0.6 $-$ distance\_factor($TU_{fl}$, {\bf $FocusList$}) \\
\> \> \> return \{[when, next($TU$$\rightarrow$$f$, $TU_{fl}$$\rightarrow$start\_date)], [certainty, $CF$]\} \\
 \\

{\em \bf Rule A4:} All cases of anaphoric relation 4. \\
for each non-empty temporal unit $TU_{fl}$ from $FocusList$ (starting with most recent) \\
\> if specificity($TU$) $\geq$ specificity($TU_{fl}$) then \\
\> \> $TU_{temp}$ = $TU_{fl}$ \\
\> \> for each \{$f$ $\vert$ $f$ $\geq$ most specific field in $TU$\} \\
\> \> \> $TU_{temp}$$\rightarrow$$f$ = null \\
\> \> if not empty merge($TU_{temp}$, $TU$) then \\
\> \> \> $CF$ = 0.5 $-$ distance\_factor($TU_{fl}$, {\bf $FocusList$}) \\
\> \> \> return \{[when, merge($TU_{temp}$, $TU$)], [certainty, $CF$]\} \\

\end{tabbing}
\caption{Main Temporal Resolution Rules}
\label{temporal-rules}
\end{figure*}
}

The algorithm does not cover a number of subcases of relations
concerning the ending times.  
For instance, rule NA2 covers only the
starting-time case of non-anaphoric relation 2. 
An example of an ending-time case that is not handled is
the utterance ``Let's meet until Thursday,''
under the meaning that they should meet
from today through Thursday.    This is an area for future
work.
\section{Results}
\label{results}
As mentioned in section \ref{intercoder}, the main results are based on
comparisons against human annotation of the held out test data. 
The results are based on straight
field-by-field comparisons of the Temporal Unit representations
introduced in section \ref{intercoder}. Thus, to be considered as correct,
information must not only be right, but it has to be in the
right place.
Thus, for example, ``Monday'' correctly resolved to Monday, 19th of
August, but incorrectly treated as a starting rather than an ending
time, contributes 3 errors of omission and 3 errors of commission (and
no credit is given for recognizing the date).

Detailed results for the test sets are presented next, starting with
results for the CMU data (see table \ref{cmu-eval}).  
Accuracy measures the degree to which the
system produces the correct answers, while precision measures the
degree to which the system's answers are correct (see the formulas
in the tables).
For each component
of the extracted temporal structure, counts were maintained for the
number of correct and incorrect cases of the system versus the tagged
file.  Since null values occur quite often, these two counts
exclude cases when one or both of the values are null. Instead,
additional counts were used for those possibilities. Note that
each test set contains three complete dialogs with an average of
72 utterances per dialog.

{
\begin{table*}
\noindent
\begin{center}
\begin{tabular}{|l|r|r|r|r|r|r|r|r|}
\hline 
Label   & Cor & Inc & Mis & Ext & Nul & AccLB &  Acc  & Prec \\ 
\hline 
start &  &  &  &  &  &  &  & \\ 
Month   &  49 &   3 &   7 &   3 &   0 & 0.338 & 0.831 & 0.891 \\ 
Date    &  48 &   4 &   7 &   3 &   0 & 0.403 & 0.814 & 0.873 \\ 
WeekDay &  46 &   6 &   7 &   3 &   0 & 0.242 & 0.780 & 0.836 \\ 
HourMin &  18 &   0 &   7 &   0 &  37 & 0.859 & 0.887 & 1.000 \\ 
TimeDay &   9 &   0 &  18 &   0 &  35 & 0.615 & 0.710 & 1.000 \\ 
\hline 
end &  &  &  &  &  &  &  & \\ 
Month   &  48 &   3 &   7 &   1 &   3 & 0.077 & 0.836 & 0.927 \\ 
Date    &  47 &   5 &   6 &   3 &   1 & 0.048 & 0.814 & 0.857 \\ 
WeekDay &  45 &   7 &   6 &   3 &   1 & 0.077 & 0.780 & 0.821 \\ 
HourMin &   9 &   0 &   9 &   0 &  44 & 0.862 & 0.855 & 1.000 \\ 
TimeDay &   4 &   0 &  13 &   1 &  44 & 0.738 & 0.787 & 0.980 \\ 
\hline 
overall & 323 &  28 &  87 &  17 & 165 & 0.428 & 0.809 & 0.916 \\ 
\hline 
\end{tabular}
\end{center}

\begin{center}
\begin{tabular}{ll}
{\bf Legend}   & \\
Cor(rect):     & System and key agree on non-null value \\
Inc(orrect):   & System and key differ on non-null value \\
Mis(sing):     & System has null value for non-null key \\
Ext(ra):       & System has non-null value for null key \\
Nul(l):        & Both System and key give null answer \\
\\
Acc(uracy)LB: & accuracy lower bound \\
Acc(uracy): & percentage of key values matched correctly \\
\multicolumn{2}{c}{(Correct + Null)/(Correct + Incorrect + Missing + Null)} \\
Prec(ision):  & percentage of System answers matching the key \\
\multicolumn{2}{c}{(Correct + Null)/(Correct + Incorrect + Extra + Null)} \\
\end{tabular}
\caption{Evaluation of System on CMU Test Data}
\label{cmu-eval}
\end{center}
\end{table*}
}

These results show that the system is performing with 81\% accuracy
overall, which is significantly better than the lower bound (defined
below) of 43\%.
In addition, the results show a high precision of 92\%.  In
some of the individual cases, however, the results could be higher due
to several factors.  For example, our system development was
inevitably focussed more on some types of slots than others.  An
obvious area for improvement is the time-of-day handling.  Also, note
that the values in the Missing column are higher than those in the
Extra column.  This reflects the conservative coding convention,
mentioned in section \ref{intercoder},  
for filling in unspecified end points.
A system that produces extraneous
values is more problematic than one that leaves entries unspecified.

Table \ref{nmsu-eval} contains the results for the system on the NMSU
data. This shows that the system performs respectably, with 69\%
accuracy and 88\% precision, on this less constrained set of data. The
precision is still comparable, but the accuracy is lower since more of
the entries were left unspecified. Furthermore, the lower bound for
accuracy (29\%) is almost 15\% lower than the one for the CMU data
(43\%), supporting the claim that this data set is more challenging.

{
\begin{table*}
\begin{center}
\noindent
\begin{tabular}{|l|r|r|r|r|r|r|r|r|}
\hline 
Label   & Cor & Inc & Mis & Ext & Nul & AccLB &  Acc  & Prec \\ 
\hline 
start &  &  &  &  &  &  &  & \\ 
Month   &  55 &   0 &  23 &   5 &   3 & 0.060 & 0.716 & 0.921 \\ 
Date    &  49 &   6 &  23 &   5 &   3 & 0.060 & 0.642 & 0.825 \\ 
WeekDay &  52 &   3 &  23 &   5 &   3 & 0.085 & 0.679 & 0.873 \\ 
HourMin &  34 &   3 &   7 &   6 &  36 & 0.852 & 0.875 & 0.886 \\ 
TimeDay &  18 &   8 &  31 &   2 &  27 & 0.354 & 0.536 & 0.818 \\ 
\hline 
end &  &  &  &  &  &  &  & \\ 
Month   &  55 &   0 &  23 &   5 &   3 & 0.060 & 0.716 & 0.921 \\ 
Date    &  49 &   6 &  23 &   5 &   3 & 0.060 & 0.642 & 0.825 \\ 
WeekDay &  52 &   3 &  23 &   5 &   3 & 0.060 & 0.679 & 0.873 \\ 
HourMin &  28 &   2 &  13 &   1 &  42 & 0.795 & 0.824 & 0.959 \\ 
TimeDay &   9 &   2 &  32 &   5 &  38 & 0.482 & 0.580 & 0.870 \\ 
\hline 
overall & 401 &  33 & 221 &  44 & 161 & 0.286 & 0.689 & 0.879 \\ 
\hline 
\end{tabular}
\caption{Evaluation of System on NMSU Test Data}
\label{nmsu-eval}
\end{center}
\end{table*}
}

More details on the lower bounds for the test data sets are shown next
(see table \ref{lower-bounds}). These values were derived by disabling
all the rules and just evaluating the input as is (after
performing normalization, so the evaluation software could be applied).
Since `null' is the
most frequent value for all the fields, this is equivalent to using a
naive algorithm that selects the most frequent value for a given
field.  The right-most column shows that there is a small
amount of error in the input representation.  This figure
is 1 minus the precision of the input representation (after
normalization).  Note, however, that this is a close but not
entirely direct measure of the error in the input, because
there are a few cases of the normalization process committing errors
and a few of it correcting them.  
Recall that the input is ambiguous; the figures in table
\ref{lower-bounds} are based on the system selecting the first ILT in
each case. Since the parser orders the ILTs based on a measure of
acceptability, this choice is likely to have the relevant temporal
information.

{
\begin{table*}
\begin{center}
\noindent
\begin{tabular}{|l|r|r|r|r|r|r|r|}
\hline
Set           & Cor & Inc & Mis & Ext & Nul &  Acc  & Input Error \\
\hline
cmu           &  84 &   6 & 360 &  10 & 190 & 0.428 & 0.055 \\
nmsu          &  65 &   3 & 587 &   4 & 171 & 0.286 & 0.029 \\
\hline
\end{tabular}
\caption{Lower Bounds for both Test Sets}
\label{lower-bounds}
\end{center}
\end{table*}
}

Since the above results are for the system taking ambiguous semantic
representations as input, the evaluation does not isolate
focus-related errors.  Therefore, two tasks were performed to aid in
developing the analysis presented in section \ref{global}.
First, anaphoric chains and competing discourse entities were manually
annotated in all of the seen data.
Second, to aid in isolating errors due to
focus issues, the system was evaluated on unambiguous, partially
corrected input for all the seen data (the test sets were retained as
unseen test data).

The overall results are shown in the table \ref{all-eval}. This
includes the results described earlier to facilitate comparisons.
Among the first, more constrained data, there are twelve dialogs in
the training data and three dialogs in a held out test set.  The
average length of each dialog is approximately 65 utterances.  Among
the second, less constrained data, there are four training dialogs
and three test dialogs.

{
\begin{table*}
\begin{center}
\begin{tabular}{|l|l|l|r|r|r|r|} 
\hline
seen/   & cmu/ & Ambiguous/  &\#dialogs &\#utterances &Accuracy &Precision \\
unseen  & nmsu & unambiguous &          &             &         &          \\
\hline
seen    & cmu & ambiguous   & 12 & 659 & 0.883 & 0.918 \\  
seen    & cmu & unambiguous & 12 & 659 & 0.914 & 0.957 \\  
unseen  & cmu & ambiguous   & 3  & 193 & 0.809 & 0.916 \\  
\hline
seen    & nmsu & ambiguous   & 4 & 358 & 0.679 & 0.746 \\  
seen    & nmsu & unambiguous & 4 & 358 & 0.779 & 0.850 \\  
unseen  & nmsu & ambiguous   & 3 & 236 & 0.689 & 0.879 \\  
\hline 
\end{tabular}

\end{center}
\caption{Results on Corrected Input (to isolate focus issues)}
\label{all-eval}
\end{table*}
}

As described in the next section, our approach
handles focus effectively.
In both data sets, there are noticeable gains in performance on the
seen data going from ambiguous to unambiguous input, especially for
the NMSU data. Therefore, the ambiguity in the dialogs contributes
much to the errors.  

The better performance on the unseen, ambiguous NMSU data over the
seen, ambiguous, NMSU data is due to several reasons. For instance,
there is vast ambiguity in the seen data. Also, numbers are mistaken by
the input parser for dates (e.g., phone numbers are treated as dates).
In addition, a tense filter, to be discussed below in section
\ref{global}, was implemented to heuristically detect subdialogs,
improving the performance of the seen NMSU ambiguous dialogs.  This
filter did not, however, significantly improve the performance for any
of the other data, suggesting that the targeted kinds of subdialogs do
not occur in the unseen data.

The errors remaining in the seen, unambiguous NMSU data are
overwhelmingly due to parser error, errors in applying the rules, 
errors in mistaking anaphoric references for deictic references
(and vice versa), and errors in choosing the wrong anaphoric relation.
As will be shown in the next section, very few errors can
be attributed to the wrong entities being in focus
due to not handling subdialogs or ``multiple threads''
(Ros\'{e} et al. 1995).
\section{Global Focus}
\label{global}
The algorithm is conspicuously lacking in any mechanism for
recognizing the global structure of the discourse, such as in
Grosz \& Sidner (1986), Mann \& Thompson (1988), Allen \& Perrault (1980), and
their descendants.  Recently in the literature, Walker (1996) has
argued for a more linear-recency based model of Attentional State
(though not that discourse structure need not be recognized), while
Ros\'{e} et al. (1995) argue for a more complex model of Attentional
State than is represented in most current computational theories of
discourse.

Many theories that address how Attentional State should be modeled
have the goal of performing intention recognition as well.
We investigate performing temporal reference resolution directly,
without also attempting to recognize discourse structure or
intentions.  We assess the challenges the
data present to our model when only this task is attempted.

We identified how far back on the focus list one must go to find
an antecedent that is appropriate according to the model.  Such
an antecedent need not be unique. (We also allow antecedents 
for which the anaphoric relation would be
a trivial extension of one of the relations in the model.)

The results are striking.  Between the two sets of data, out of 215
anaphoric references, there are fewer than 5\% for which the
immediately preceding time is not an appropriate antecedent.  Going
back an additional time covers the remaining cases.

The model is geared toward allowing the most recent Temporal Unit to
be an appropriate antecedent.  For example, in the example for
anaphoric relation \ref{revise}, the second utterance (as well as
the first) is a possible antecedent of the third.  A corresponding
speech act analysis might be that the speaker is suggesting a
modification of a previous suggestion.  Considering the most recent
antecedent as often as possible supports robustness, in the sense that
more of the dialog is considered.

There are subdialogs in the NMSU data (but none in the CMU data) for
which our recency algorithm fails because it lacks a mechanism for
recognizing subdialogs.  There are five temporal references within
subdialogs that recency either incorrectly interprets 
to be anaphoric to a time mentioned before the subdialog or incorrectly 
interprets to be
the antecedent of a time mentioned after the subdialog. 
Fewer than 25 cumulative errors result from these primary areas.  In
the case of one of the primary errors,
recency commits a
``self-correcting'' error; without this luck, the remainder of the
dialog would have represented additional cumulative error.

In a departure from the algorithm, the system uses a simple heuristic
for ignoring subdialogs:
a time is ignored if the utterance evoking it is in the simple past
or past perfect.   This prevents a number of the above errors
and suggests that 
changes in tense, aspect, and modality are promising clues to explore for
recognizing subdialogs in this kind of data (cf., e.g., 
Grosz \& Sidner 1986;
Nakhimovsky 1988).   The CMU data has very little variation
in tense and aspect, the reason a mechanism for interpreting
them was not incorporated into the algorithm.  

Ros\'e et al. (1995) report that ``multiple threads'', when the
participants are negotiating separate times,
pose challenges to a stack-based discourse model on both the
intentional and attentional levels.  They posit a more complex
representation of Attentional State to meet these challenges.
They report improved results on speech-act resolution
in a corpus of scheduling dialogs.

Here, we focus on just the attentional level.  The structure relevant
for the task addressed in this paper is the following,
corresponding to their figure 2.
There are four Temporal Units mentioned in the order $TU_1$, $TU_2$,
$TU_3$, $TU_4$ (other times could be mentioned in between).  The
(attentional) multiple thread case is when $TU_1$ is required to be an
antecedent of $TU_3$, but $TU_2$ is also needed to interpret $TU_4$.
Thus, $TU_2$ cannot be simply thrown away or ignored once we are done
interpreting $TU_3$.  This structure would definitely pose a difficult
problem for our algorithm, but there are no realizations, in terms of
our model, of this structure in the data we analyzed.

The different findings might be due to the fact that different
problems are being addressed.  Having no intentional state, our model
does not distinguish times being negotiated from other times.  It is
possible that another structure is relevant for the intentional level:
Ros\'{e} et al. (1995) do not specify whether or not this is so.  The
different findings may also be due to differences in the data: although
their scheduling dialogs were collected under similar
protocols, their protocol is like a radio conversation in which a
button must be pressed in order to transmit, resulting in less
dynamic interaction and longer turns (Villa~1994).

An important discourse feature of the dialogs  is the degree of redundancy
of the times mentioned (Walker 1996). This limits the ambiguity of the times
specified, and it also leads to a higher level of robustness, since
additional DE's with the same time are placed on the focus
list. These ``backup'' DE's might be available in case the rule
applications fail on the most recent DE.  Table \ref{redundancy}
presents measures of redundancy.
For illustration, the redundancy is broken down into the case
where redundant plus additional information is provided 
(``redundant'') versus the case where the temporal information
is just repeated (``reiteration''). This shows that
roughly 25\% of the CMU utterances with temporal information contain
redundant temporal references, while 20\% of the NMSU ones do.

\begin{table*}
\begin{center}
\begin{tabular}{|c|c|c|c|c|} 
\hline 
Dialog Set & Temporal Utterances & Redundant & Reiteration & \% \\ 
\hline 
cmu     & 210            & 36        & 20          & 26.7 \\ 
nmsu    & 122            & 11        & 13          & 19.7 \\ 
\hline
\end{tabular}
\end{center}
\caption{Redundancy in the Training Dialogs}
\label{redundancy}
\end{table*}

\section{Conclusions}
This paper presented an intercoder reliability study showing strong
reliability in coding the temporal information targeted in this work.
A model of temporal reference resolution in scheduling dialogs was
presented which supports linear recency and has very good coverage;
and, an algorithm based on the model was described.
The analysis of the detailed results showed that the implemented
system performs quite well (for instance, 81\% accuracy vs. a lower
bound of 43\% on the unseen CMU test data).

We also assessed the challenges presented by the data to a method that does
not recognize discourse structure, based on an extensively annotated
corpus and our experience developing a fully automatic system.  In an
overwhelming number of cases, the last mentioned time is an appropriate
antecedent with respect to our model, in both the more and the less
constrained data.  In the less constrained data, some error occurs due
to subdialogs, so an extension to the approach is needed to handle
them.  But in none of these cases would subsequent errors result if,
upon exiting the subdialog, the offending information were popped off
a discourse stack or otherwise made inaccessible.  Changes in tense,
aspect, and modality are promising clues for recognizing subdialogs in
this data, which we plan to explore in future work.
\section{Acknowledgements}
This research was supported in part by the Department of
Defense under grant number 0-94-10.   
A number of people contributed to this work.  We want to
especially thank David Farwell, Daniel Villa, Carol Van Ess-Dykema, 
Karen Payne,
Robert Sinclair, Rocio Guill\'{e}n,  David Zarazua,
Rebecca Bruce, Gezina Stein, Tom Herndon, and
CMU's Enthusiast project members, 
whose cooperation greatly aided our project.

\bibliographystyle{plain}

\end{document}